\documentclass[final]{cvpr}

\usepackage{times}
\usepackage{epsfig}
\usepackage{graphicx}
\usepackage{amsmath}
\usepackage{amssymb}
\usepackage{xcolor}
\usepackage{tabularx}
\usepackage{bbm}
\usepackage{capt-of}
\usepackage{adjustbox}

\definecolor{myred}{RGB}{190,20,20}
\usepackage[pagebackref=true,breaklinks=true,colorlinks,bookmarks=false]{hyperref}

\DeclareMathOperator*{\argmin}{arg\,min}

\usepackage[normalem]{ulem}
\usepackage{ifthen}

\usepackage{xspace}
\usepackage{enumitem}
\usepackage[leftmargin=2mm,rightmargin=0pt]{quoting}
\newcommand{\mathcls}{[\mathsf{CLS}]}
\newcommand{\cls}{$\mathcls{}$\xspace}

\newcommand{\dinovit}{DINO-ViT\xspace}

\newcommand{\myparagraph}[1]{\vspace{0.15cm}\noindent{\bf #1}\hspace{0.05cm}} 

\def\confYear{CVPR 2022}

\newcommand{\afterfigure}{\vspace{-1em}}

\begin{document}

\title{Splicing ViT Features for Semantic Appearance Transfer}
\author{Narek Tumanyan$^*$\qquad Omer Bar-Tal$^*$\qquad Shai Bagon \qquad Tali Dekel \\
{\small Weizmann AI Center (WAIC), Dept. of Computer Science and Applied Math, The Weizmann Inst. of Science} \\
{\small *Indicates equal contribution.} \\
{\small Project webpage: \url{https://splice-vit.github.io}}
}
\twocolumn[{
\renewcommand\twocolumn[1][]{#1}
\maketitle
\centering
\vspace*{-8mm}
\includegraphics[width=1\textwidth]{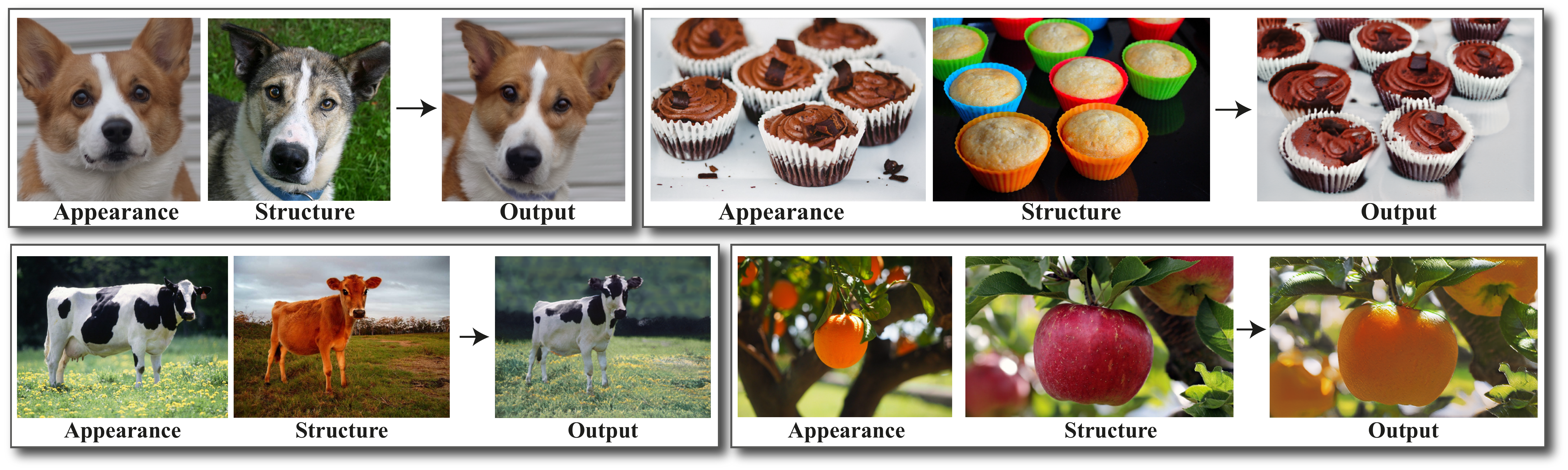}\vspace{-0.1cm}
\captionof{figure}{Given two input images---a source \emph{structure} image and a target \emph{appearance} image--our method generates a new image in which the structure of the source image is preserved, while the visual appearance  of the target image is transferred in a \emph{semantically} aware manner. 
That is, objects in the structure image are ``painted'' with the visual appearance of semantically related objects in the appearance image. Our method leverages a self-supervised, pre-trained ViT model as an external semantic prior. This allows us to train our generator only on a single input image pair, without any additional information (e.g., segmentation/correspondences), and without adversarial training. Thus, our framework can work across a variety of objects and scenes, and can generate high quality results in high resolution (e.g., HD).
}\label{fig:teaser}
 \vspace*{0.26cm}
}]


\maketitle
\begin{abstract}
\vspace*{-1.05em}
We present a method for semantically transferring the visual appearance of one natural image to another. 
Specifically, our goal is to generate an image in which objects in a source structure image are ``painted'' with the visual appearance of their semantically related objects in a target appearance image.  Our method works by training a generator given only a single  structure/appearance image pair as input. To integrate semantic information into our framework---a pivotal component in  tackling this task---our key idea is to leverage a pre-trained and fixed Vision Transformer (ViT) model which serves as an external semantic prior.  Specifically, we derive novel representations of structure and  appearance extracted from deep ViT features, untwisting them from the learned self-attention modules. We then establish an objective function that splices the desired  structure and appearance representations, interweaving them together in the space of ViT features. 
Our framework, which we term ``Splice", does not involve adversarial training, nor does it require any additional input information such as semantic segmentation or correspondences, and can generate high resolution results, e.g., work in HD. We demonstrate high quality results on a variety of in-the-wild image pairs, under significant variations in the number of objects, their pose and appearance.
\end{abstract}

\section{Introduction}
\begin{quoting}
\noindent\textit{``Rope splicing is the forming of a semi-permanent joint between two ropes by partly untwisting and then interweaving their strands.''~\cite{splicing}} 
\end{quoting}
\vspace{-0.1cm}What is required to transfer the visual appearance between two semantically related 
images? 
Consider for example the task of transferring the visual appearance of a spotted cow in a flower field to an image of a red cow in a grass field (Fig.~\ref{fig:teaser}). Conceptually, we have to associate regions in both images that are semantically related, and transfer the visual appearance between these matching regions. Additionally, the target appearance has to be transferred in a realistic manner, while preserving the structure of the source image~-- the red cow should be realistically ''painted'' with black and white spots, and the green grass should be covered with yellowish colors.  To achieve it under noticeable pose, appearance and shape differences between the two images, \emph{semantic} information is imperative.

Indeed, with the rise of deep learning and the ability to learn high-level visual representations from data, new vision tasks and methods under the umbrella of ``visual appearance transfer'' have emerged. For example, the image-to-image translation trend aims at translating a source image from one domain to another target \emph{domain}. To achieve that, most methods use generative adversarial networks (GANs), given image collections from both domains. Our goal is different --  rather than generating \emph{some} image in a target domain, we generate an image that depicts the visual appearance of a \emph{particular} target image, while preserving the structure of the source image. Furthermore, our method is trained using only a single image pair as input, which allows us to deal with scenes and objects for which an image collection from each domain is not handy (e.g., spotted cows and red cows image collections). 

With only a pair of images available as input, how can we source semantic information? We draw inspiration from Neural Style Transfer (NST) that represents content and an artistic style in the space of deep features encoded by a pre-trained classification CNN model (e.g., VGG). While NST methods have shown a remarkable ability to \emph{globally} transfer  artistic styles, their content/style representations are not suitable for \emph{region-based}, semantic appearance transfer across objects in two natural images~\cite{JingYFYYS20}.  Here, we propose novel deep representations of  appearance and structure that are extracted from DINO-ViT~-- a Vision Transformer model that has been pre-trained in a self-supervised manner~\cite{dino}. Representing structure and appearance in the space of ViT features allows us to inject powerful semantic information into our method and establish a novel objective function  that is used to train a generator using only the single input image pair. 

DINO-ViT has been shown to learn powerful and meaningful visual representation, demonstrating impressive results on several downstream tasks including image retrieval, object segmentation, and copy detection~\cite{dino, amir2021deep}. However, the intermediate representations that it learns have not yet been fully explored.  We thus first strive to gain a better understanding of the information encoded in different ViT's features across layers. We do so by adopting  ``feature inversion'' visualization techniques previously used in the context of CNN features. Our study provides a couple of key observations: (i)~the global token (a.k.a \cls token) provides a powerful representation of visual appearance, which captures not only texture information but more global information such as object parts, and (ii) the original image can be reconstructed from these features, yet they provide powerful semantic information at high spatial granularity.

Equipped with the above observations,  we derive novel representations of structure and visual appearance extracted from deep ViT features -- untwisting them from the learned self-attention modules. Specifically, we represent visual appearance via   the  global  \cls  token,  and represent structure via the self-similarity of keys, all extracted from the last layer. We then train a generator on a single input pair of structure/appearance images, to produce an image that \emph{splices} the desired visual appearance and structure in the space of ViT features.  Our framework does not require any additional information such as semantic segmentation and does not involve adversarial training. Furthermore, our model can be trained on high resolution images, producing high quality results in HD. We demonstrate a variety of semantic appearance transfer results across diverse natural image pairs, containing significant variations in the number of objects, pose and appearance.

\section{Related Work} \label{sec:related}
The problem we tackle here is  \emph{semantic} visual appearance transfer between two \emph{in-the-wild}, {\em natural} images, without user guidance. To the best of our knowledge, there is no existing method addressing  specifically this challenge. We  review the most related trends and methods. 

\myparagraph{Domain Transfer \& Image-to-Image Translation.} The goal of these methods is to learn a mapping between source and target \emph{domains}.
This is typically done by training a GAN on a \emph{collection} of images from the two domains,
either paired~\cite{8100115} or unpaired~\cite{8237506, 10.5555/3294771.3294838, kim2017learning, 8237572, park2020contrastive}.
Swapping Autoencoder (SA)~\cite{park2020swapping} trains a domain-specific GAN to disentangle structure and texture in images,
and swap these representations between two images in the domain.
In contrast to SA, our method  is not restricted to any particular domain, and it does not require a collection of images for training, nor it involves adversarial training.

Recently, image to image translation methods trained on a single example were proposed~\cite{cohen2019bidirectional, DBLP:journals/cgf/BenaimMBW21, lin2020tuigan}. 
These methods only utilize low-level visual information and lack semantic understanding.
Our method is also trained only on a single image pair, but leverages a pretrained ViT model to inject powerful semantic information into the generation process (see Sec.~\ref{sec:results} for comparison).

\myparagraph{Neural Style Transfer (NST).}
In its classical setting, NST transfers an \emph{artistic} style from one image to another~\cite{GatysEBHS17,JingYFYYS20}.
STROTSS~\cite{kolkin2019style} uses pre-trained VGG features to represent style and their self-similarity to capture structure in an optimization-based framework to perform \emph{artistic} style transfer in a \emph{global} manner.
In contrast, our goal is to  transfer the appearance between \emph{semantically} related objects and regions in two \emph{natural} images. 

\emph{Semantic style transfer} methods also aim at mapping appearance across semantically related regions between two images~\cite{MechrezTZ18, LiW16, Wilmot2017StableAC, Fast_Photographic}. 
However, these methods are usually restricted to color transformation~\cite{XuWFSZ20, Fast_Photographic, Yoo_2019_ICCV}, or depend on additional semantic inputs (e.g., annotations, segmentation, point correspondences, etc.)~\cite{GatysEBHS17, kim2020deformable, Champandard16, kolkin2019style}.  Other works tackle the problem for specific controlled domains~\cite{BarnesF14a,ShihPDF13}.
In contrast, we aim to work with arbitrary, in-the-wild input pairs.

\myparagraph{Vision Transformers (ViT).}
ViTs~ \cite{vit} have been shown to achieve competitive results to state-of-the-art CNN architectures on image classification tasks, while demonstrating impressive robustness
\cite{naseer2021intriguing}.
\dinovit~\cite{dino} is a ViT model that has been trained, without labels, using a self-distillation approach.
The effectiveness of the learned representation  has been demonstrated 
on several downstream tasks, including image retrieval and segmentation. 

\emph{Amir et al.}~\cite{amir2021deep}
have demonstrated the power of \dinovit Features as dense visual descriptors. Their key observation is that deep DINO-ViT features  capture rich semantic information at fine spatial granularity, e.g, describing semantic object \emph{parts}. Furthermore, they observed that the representation is shared across different yet related object classes. 
This power of \dinovit features was exemplified by performing ``out-of-the-box" unsupervised semantic part co-segmentation and establishing semantic correspondences across different objects categories. 
Inspired by these observations, we harness the power of \dinovit features in a novel generative direction -- we derive new perceptual losses capable of splicing structure and semantic appearance across semantically related objects.

\section{Method}

\begin{figure}[t!]
    \centering
    \includegraphics[width=.5\textwidth]{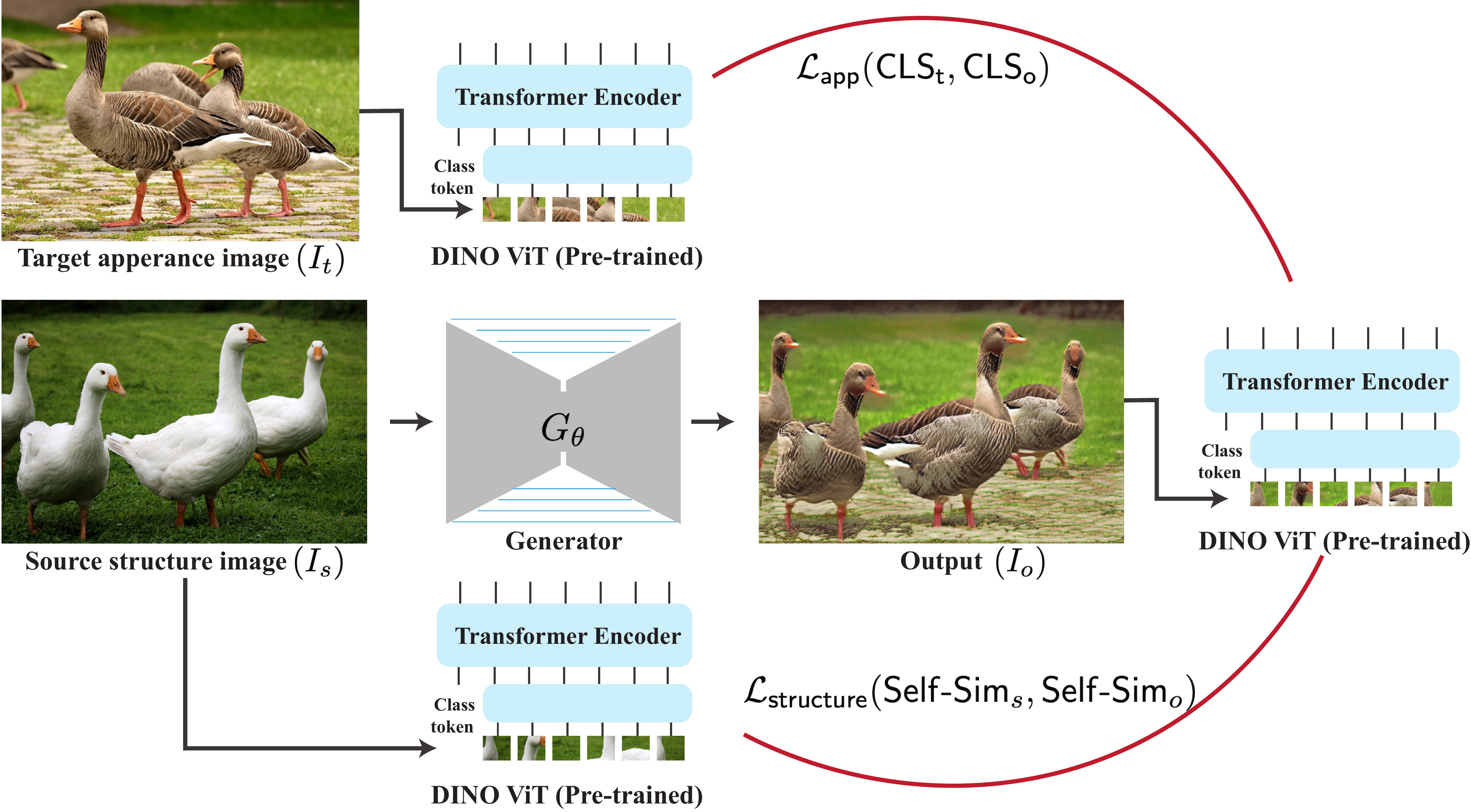}
    \caption{{\bf Pipeline.} Our generator $G_\theta$  takes an input structure image $I_s$ and outputs $I_o$. We establish our training losses using a pre-trained and fixed DINO-ViT model, which serves as an external semantic prior: we represent \emph{structure} via the self-similarity  of keys in the deepest attention module ($\mathsf{Self}\text{-}\mathsf{Sim}$), and \emph{appearance} via the \cls token in the deepest layer. Our objective is twofold: (i)   $\mathcal{L}_{\mathsf{app}}$ encourages the \cls of $I_o$ to match the \cls of $I_{t}$, and (ii) $\mathcal{L}_{\mathsf{structure}}$  encourages the self-similarity representation of $I_o$ and  $I_s$ to be the same. See Sec.~\ref{sec:method} for details.
    }
    \label{fig:pipeline} \afterfigure
\end{figure}
Given a source structure image $I_s$ and a target appearance image $I_t$, our goal is to generate an image $I_o$, in which objects in  $I_s$ are ``painted'' with the visual appearance of their semantically related objects in $I_t$.

Our framework is illustrated in Fig.~\ref{fig:pipeline}: for a given pair $\{I_s, I_t\}$, we train a generator $G_\theta(I_s)=I_o$. To establish our training losses, we leverage \dinovit~-- a  self-supervised, pre-trained ViT model~\cite{dino}~-- which is kept fixed and serves as an external high-level prior.  We propose new deep representations for \emph{structure} and \emph{appearance} in DINO-ViT feature space; we  train $G_\theta$ to output an image, that when fed into DINO-ViT, matches the source structure and target appearance representations.
Specifically, our training objective is twofold: (i) $\mathcal{L}_{\mathsf{app}}$ that encourages the deep appearance  of $I_o$ and $I_t$ to match, and (ii) $\mathcal{L}_{\mathsf{structure}}$, which encourages the deep structure  representation of $I_o$ and  $I_s$ to match.

We next  briefly review ViT architecture, then provide qualitative analysis of \dinovit's features in Sec.~\ref{sec:inversion}, and describe our framework  in Sec.~\ref{sec:method}.  

\subsection{Vision Transformers -- overview}
\label{sec:vit}

In ViT, an image $I$ is processed as a sequence of $n$ non-overlapping patches as follows: first, \emph{spatial tokens} are formed by linearly embedding each patch to a $d$-dimensional vector, and adding learned position embeddings. An additional learnable token,  a.k.a \cls token,  serves as a global representation of the image.

The set of tokens are then passed through $L$ Transformer layers, each consists of normalization layers (LN), Multihead Self-Attention (MSA) modules, and MLP blocks:
{\small \begin{equation*}
\begin{array}{l}
     \hat{T}^{l} = \mathsf{MSA}(\mathsf{LN}(T^{l-1}))  + T^{l-1}\\ 
     T^{l} = \mathsf{MLP}(\mathsf{LN}(\hat{T}^{l})) + \hat{T}^{l}
\end{array}
\end{equation*}}
where $T^{l}(I)\!=\!\left[t_{\textit{cls}}^{l}(I), t_1^l(I) \dots t_n^{l}(I)\right]$ are the output tokens for layer $l$ for image $I$.

In each MSA block
the (normalized) tokens are linearly projected into queries, keys and values:. 
{\small \begin{equation}\label{eq:qkv}
    Q^l = T^{l-1}\cdot W_{q}^l,\; K^l = T^{l-1}\cdot W_{k}^l ,\; V^l_h = T^{l-1}\cdot W_{v}^l
\end{equation}}
which are then fused using multihead self-attention to form the output of the MSA block (for full details see~\cite{vit}).

After the last layer, the \cls token is passed through an additional MLP to form the final output, e.g., output distribution over  a  set of labels \cite{vit}.
In our framework, we leverage \dinovit~\cite{dino}, in which the model has been trained in a self-supervised manner using a self-distillation approach. 
Generally speaking,  the model is trained to produce  the same  distribution  for  two different augmented views of the same image. As  shown in \cite{dino}, and in~\cite{amir2021deep}, \dinovit learns  powerful visual representations that are less noisy and more semantically meaningful than the supervised ViT.

\begin{figure*}[t!]
    \centering
    \includegraphics[width=\textwidth]{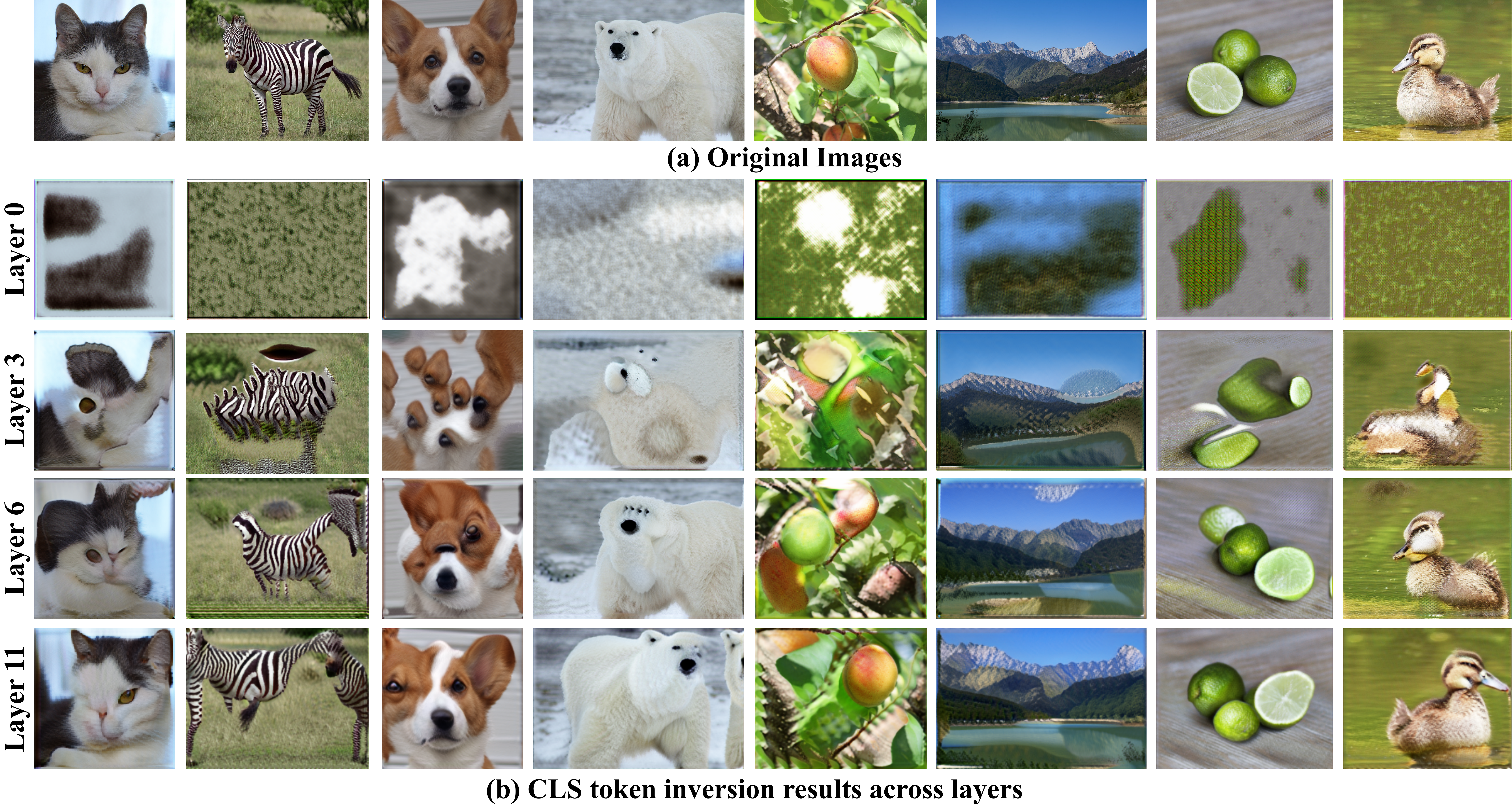}
    \caption{{\bf Inverting the [CLS] token across layers.} Each input image (a) is fed to \dinovit to compute its global [CLS] token at different layers. (b) Inversion results: starting from a noise image, we optimize for an image that would match the original [CLS] token at a specific layer. While earlier layers capture local texture, higher level information such as object parts emerges at the deeper layers (see Sec.~\ref{sec:inversion}). }\afterfigure
    \label{fig:inversion}
\end{figure*}

\subsection{Structure \& Appearance in ViT's Feature Space}
\label{sec:inversion}

The pillar of our method is the representation  of \emph{appearance} and  \emph{structure} in the space of \dinovit features. 
For  appearance, we want a representation that can be spatially flexible, i.e., discards the exact objects' pose and scene's spatial layout, while capturing global appearance information and style. To this end, we leverage the \cls token, which serves as a \emph{global} image representation.

For structure, we want a representation that is robust to local texture patterns, yet preserves the spatial layout, shape and perceived semantics of the objects and their surrounding. 
To this end, we leverage  deep \emph{spatial} features extracted from \dinovit, and use their \emph{self-similarity} as structure representation:

{\small \begin{equation}
    S^L(I)_{ij} = \mathsf{cos}\text{-}\mathsf{sim}\left( k^{L}_{i}(I), k^{L}_{j}(I)\right)  \label{eq:selfsim}
\end{equation}}
$\mathsf{cos}\text{-}\mathsf{sim}$ is the cosine similarity between keys (See Eq.~\ref{eq:qkv}).
Thus, the dimensionality of our self-similarity descriptor becomes $S^{L}(I) \in \mathbb{R}^{(n+1)\!\times\!(n+1)}$, where $n$ is the number of patches.

The effectiveness of self-similarly-based descriptors in capturing \emph{structure} while ignoring \emph{appearance} information have been previously demonstrated by both classical methods~\cite{shechtman2007localselfsim}, 
and recently also using deep CNN features for artistic style transfer~\cite{kolkin2019style}. 
We opt to use the self similarities of \emph{keys}, rather than other facets of ViT, based on~\cite{amir2021deep}. 

\begin{figure}
    \centering
    \includegraphics[width=.45\textwidth]{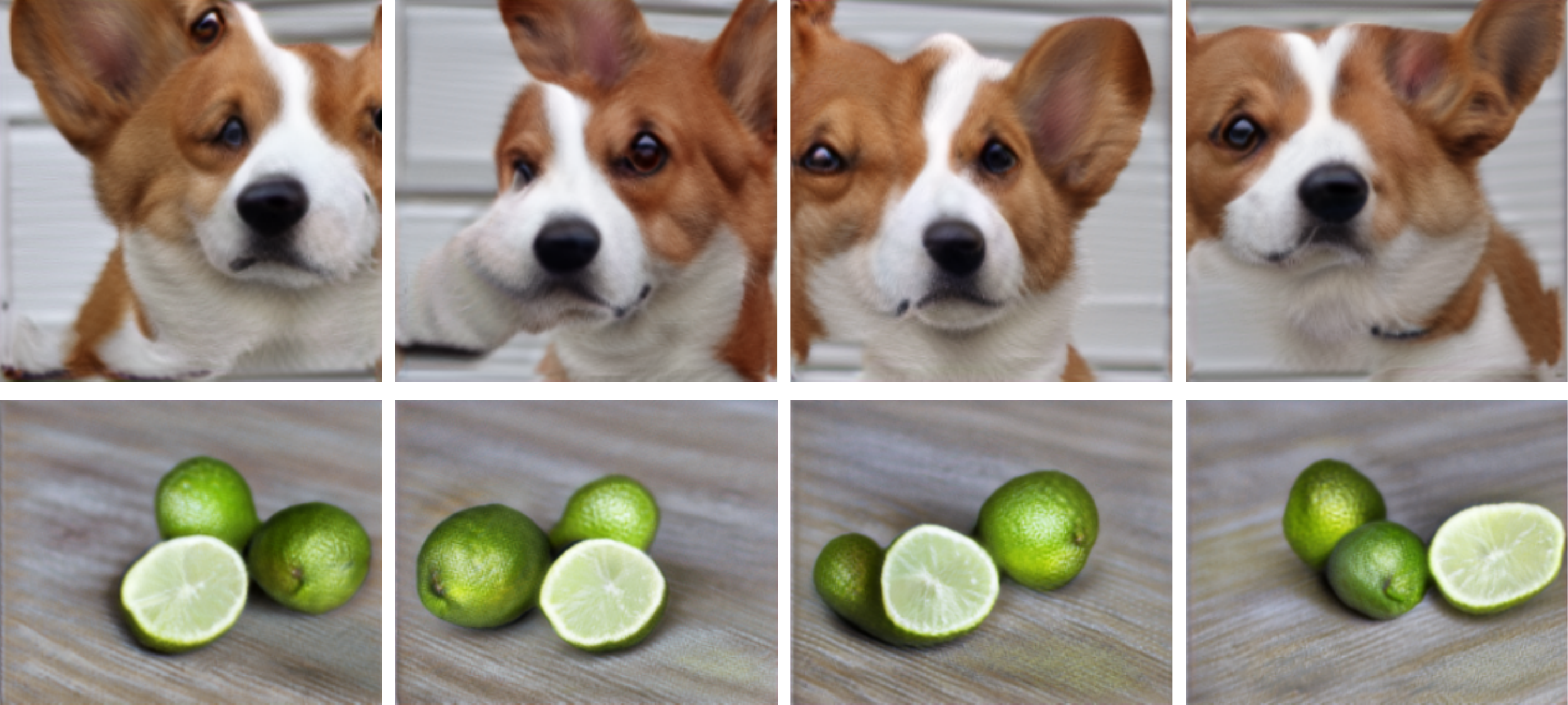}
    \caption{{\bf [CLS] token inversion over multiple runs.} The variations in structure in multiple inversion runs of the same image demonstrates the spatial flexibility of the [CLS] token.}
    \label{fig:inv_runs}\afterfigure
\end{figure}

\myparagraph{Understanding and visualizing DINO-ViT's features.} To better understand our ViT-based representations, we take a \emph{feature inversion} approach~-- given an image, we extract target features, and optimize for an image that has the same features. Feature inversion has been widely explored in the context of CNNs (e.g., \cite{simonyan2014deep,mahendran2014understanding}), however has not been attempted for understanding ViT features yet. For CNNs, it is well-known that solely optimizing the image pixels is insufficient for converging into a meaningful result~\cite{olah2017feature}. We observed a similar phenomenon when inverting ViT features (see Supplementary Materials on our website~--~SM). Hence, we incorporate ``Deep Image Prior`` \cite{UlyanovVL17}, i.e., we optimize for the weights of a CNN  $F_\theta$ that translates a fixed random noise $z$ to an output image:
{\small \begin{equation}
    \argmin_\theta ||\phi(F_\theta(z)) - \phi(I)||_F,
    \label{eq:inversion}
\end{equation}}
where $\phi(I)$ denotes the target features, and $|| \cdot ||_F$ denotes Frobenius norm.
First, we consider inverting the \cls token: $\phi(I)= t_{\textit{cls}}^l(I)$. Figure~\ref{fig:inversion} shows our inversion results across layers, which illustrate the following observations: 
\begin{enumerate}[leftmargin=*]
    \item From shallow to deep layers, the \cls token gradually  accumulates appearance  information. Earlier layers mostly capture local texture patterns, while in deeper layers, more global information such as object parts  emerges.
    \item The \cls token encodes  appearance information in a \emph{spatially flexible manner}, i.e., different object parts can stretch, deform or  be flipped.  Figure~\ref{fig:inv_runs} shows multiple runs of our inversions per image; in all runs, we can notice similar global information, but the diversity across runs demonstrates the spatial flexibility of the representation.
\end{enumerate}

Next, in Fig.~\ref{fig:keys}(a), we show the inversion of the spatial keys extracted from the last layer, i.e., $\phi(I) = K^{L}(I)$. These features have been shown to encode high level information~\cite{dino, amir2021deep}. Surprisingly, we observe that the original image can still be  reconstructed from this representation.

To discard appearance information encoded in the keys, we consider the self-similarity of the keys (see Sec.~\ref{sec:inversion}). 
This is demonstrated in the PCA visualization of the keys' self-similarity in Fig.~\ref{fig:keys}(b). As seen, the self-similarity mostly captures  the  structure of objects, as well as their distinct semantic components. For example, the legs and the body of the polar bear that have the same texture, are distinctive.

\subsection{Splicing ViT Features}
\label{sec:method}

Based on our understanding of \dinovit's internal representation, we turn to the task of training our generator. 

Our objective function takes the following form:

{\small \begin{equation} \label{eq:4}
    \mathcal{L}_{\mathsf{splice}} =   \mathcal{L}_{\mathsf{app}} + \alpha \mathcal{L}_{\mathsf{structure}} + \beta \mathcal{L}_{\mathsf{id}},
\end{equation}}
where $\alpha$ and $\beta$ set the relative weights between the terms. The driving loss of our objective function is $\mathcal{L}_{\mathsf{app}}$, and we set
 $\alpha=0.1, \beta=0.1$ for all experiments.
 
\myparagraph{Appearance loss.} The term $\mathcal{L}_{\mathsf{app.}}$ encourages the output image to match the appearance of $I_t$, and is defined as the difference in \cls  token between the generated and texture image: 
{\small
\begin{equation} \label{eq:6}
    \mathcal{L}_{\mathsf{app}} = \left\| t_{\mathcls}^{L}(I_t) - t_{\mathcls}^{L}(I_o) \right\|_2 ,
\end{equation}}
where $t_{\mathcls}^{L}(\cdot) = t_{cls}^{L}$ is the \cls token extracted from the deepest layer (see Sec.~\ref{sec:vit}).

\myparagraph{Structure loss.} The term $\mathcal{L}_{\mathsf{structure}}$ encourages the output image to match the structure of $I_s$, and is defined by the difference in self-similarity of the keys extracted from the attention module at deepest transformer layer: 

\begin{figure}[t!]
    \centering
    \includegraphics[width=.5\textwidth]{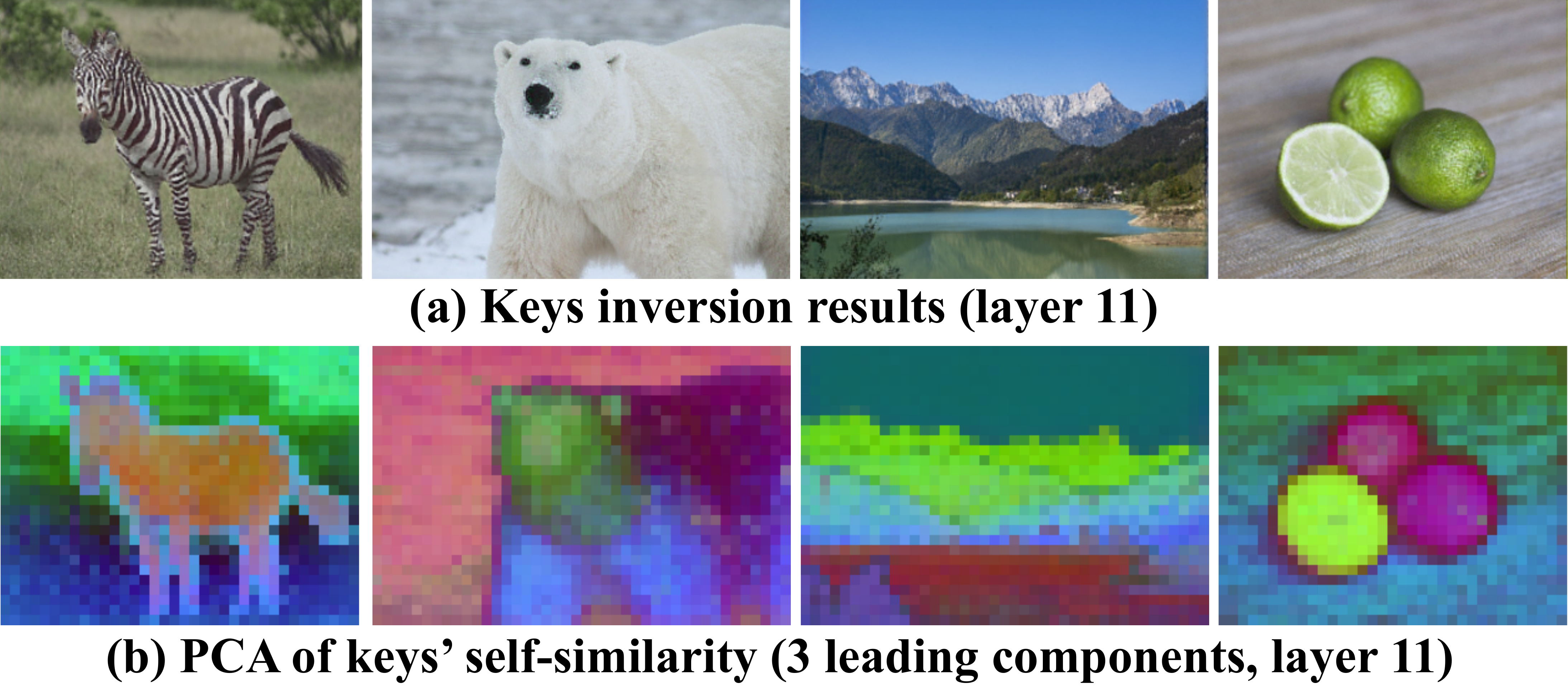}
    \caption{{\bf Visualization of DINO-ViT keys.} (a) Inverting keys from the deepest layer surprisingly reveals  that the  image can be reconstructed.  (b) PCA visualization of the  keys' self-similarity: the leading components mostly  capture semantic scene/objects parts, while discarding appearance information (e.g., zebra stripes). }\afterfigure
    \label{fig:keys}
\end{figure}

{\small \begin{equation} \label{eq:5}
    \mathcal{L}_{\mathsf{structure}} = \left\|S^{L}(I_s) - S^{L}(I_o) \right\|_F,
\end{equation}}
where $S^{L}(I)$ is defined in Eq.~(\ref{eq:selfsim}).
\begin{figure*}
    \centering
    \includegraphics[width=\textwidth]{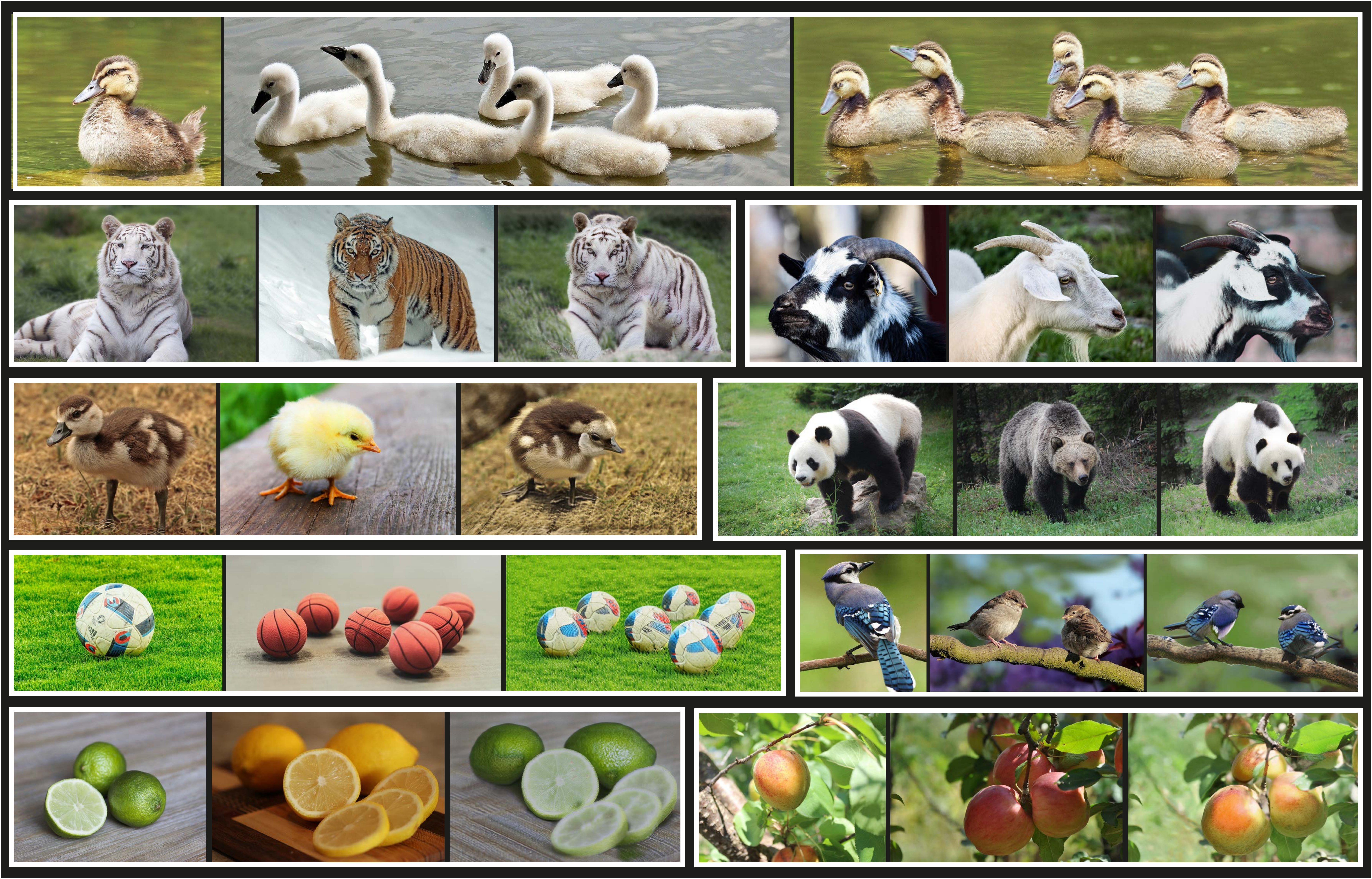}
    \caption{{\bf Sample results on in-the-wild image pairs.} For each example, shown left-to-right: the target appearance image, the source structure image and our result. The full set of results is included in the SM. Notice the variability in number of objects, pose, and the significant appearance changes between the images in each pair. } \afterfigure
    \label{fig:results}
\end{figure*}

\myparagraph{Identity Loss.} The term $\mathcal{L}_{\mathsf{id}}$ is used as a regularization. Specifically, when we feed $I_t$ to the generator, this loss encourages $G_\theta$ to preserve the keys representation of $I_t$: 
{\small \begin{equation} \label{eq:7}
    \mathcal{L}_{{\mathsf{id}}} = \left\| K^{L}(I_t) - K^{L}(G_\theta(I_t)) \right\|_F \\
\end{equation}}

Similar loss terms, defined in RGB space, have been used as a regularization in training GAN-based generators for image-to-image translation \cite{park2020contrastive, DBLP:conf/iclr/TaigmanPW17, 8237506}. Here, we apply the identity loss with respect to the \emph{keys} in the deepest ViT layer, a semantic yet invertible representation of the input image (as discussed in section \ref{sec:inversion}).

\myparagraph{Data augmentations and training.} Since we only have a single input pair $\{I_s, I_t\}$, we create additional training examples, $\{I^i_s, I^i_t\}_{i=1}^N$, by applying augmentations such as crops and color jittering (see Appendix~\ref{sec:appendix-aug} for implementation details).  $G_\theta$ is now trained  on multiple \emph{internal examples}. Thus, it has to learn a good mapping function for a \emph{dataset} containing $N$ examples, rather than solving a test-time optimization problem for a single instance.  Specifically, for each example, the objective is to generate $I_o^i = G_\theta(I_s^i)$, that matches the structure of $I_s^i$ and the appearance of $I_t^i$.

\begin{figure*}[t!]
    \centering
    \includegraphics[width=1\textwidth]{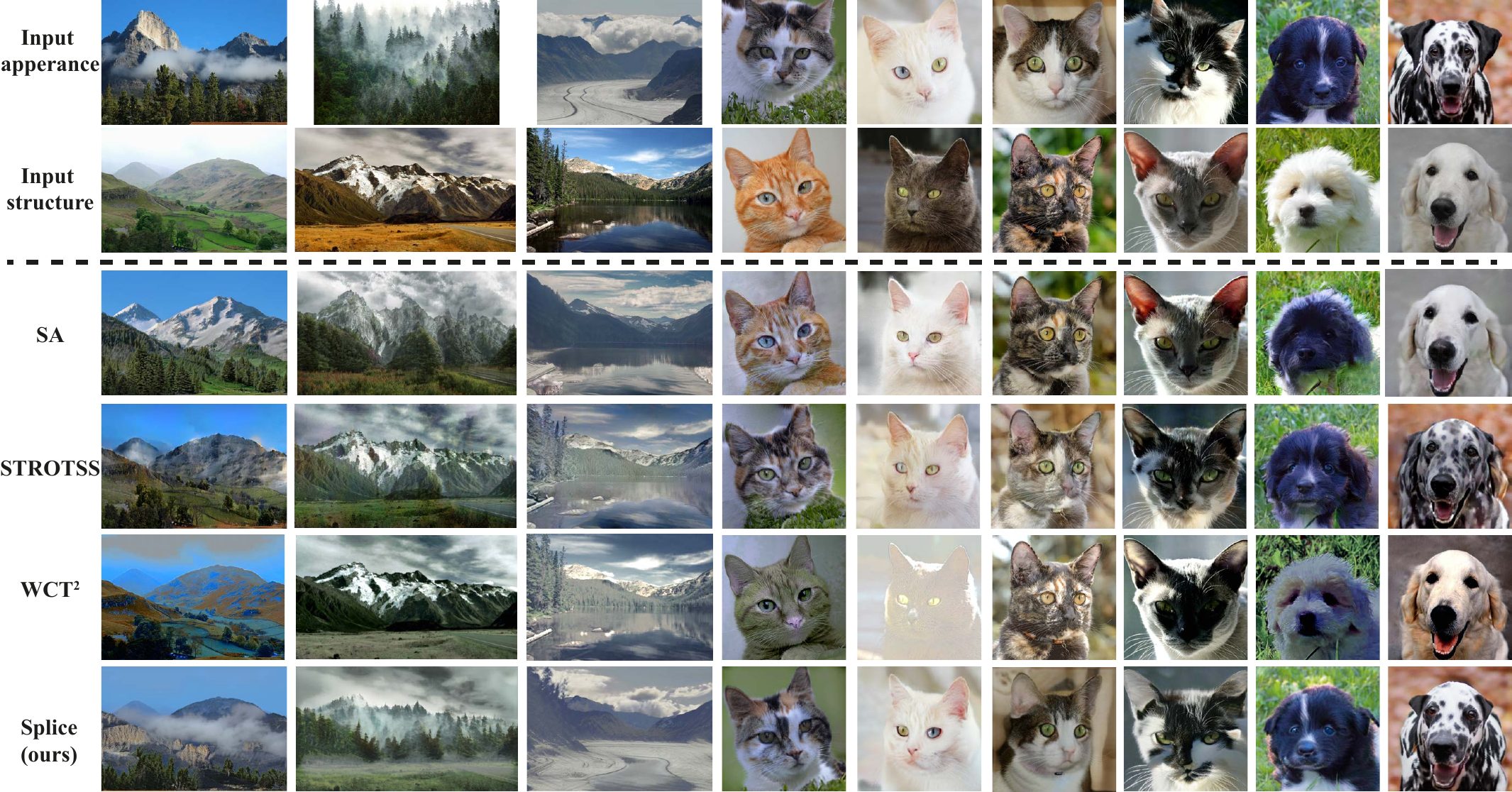}
    \caption{{\bf Comparisons with style transfer and swapping autoencoders.} First two rows: input appearance and structure images taken from the AFHQ and Flickr Mountains. The following rows, from top to bottom, show the results of: swapping autoencoders (SA) \cite{park2020swapping},  STROTSS \cite{kolkin2019style}, and  $\text{WCT}^2$ \cite{Yoo_2019_ICCV}.  See SM for additional comparisons.}\afterfigure
    \label{fig:comparisons}
\end{figure*}

\section{Results}
\label{sec:results}
\myparagraph{Datasets.} We tested our method on a variety of image pairs gathered from Animal Faces
HQ (AFHQ) dataset \cite{Choi_2020_CVPR}, and images crawled from \href{https://flickr.com/groups/justmountains/pool/}{Flickr Mountain}. In addition, we collected our own dataset, named \emph{Wild-Pairs}, which includes a set of 25 high resolution image pairs taken from \href{https://pixabay.com}{Pixabay}, each pair depicts semantically related objects from different categories including animals, fruits, and other objects.  The number of objects, pose and appearance may significantly change between the images in each pair.  The image resolution ranges from 512px to 2000px.

Sample pairs from our dataset along with our results can be seen in Fig.~\ref{fig:teaser} and Fig.~\ref{fig:results}, and the full set of pairs and results is included in the SM. As can be seen, in all examples, our method  successfully transfers the visual appearance in a semantically meaningful manner at several levels:  (i) \emph{across objects:} the target visual appearance of objects is being transferred to to their semantically related objects in the source structure image, under significant variations in pose, number of objects, and appearance between the input images.  (ii) \emph{within objects:} visual appearance is transferred between corresponding body parts or object elements. For example, in Fig.~\ref{fig:results} top row, we can see the appearance of a single duck is semantically transferred to each of the 5 ducks in the source image, and that the appearance of each body part is mapped to its corresponding part in the output image.  This can be consistently observed in all our results. 

The results  demonstrate that our method is capable of performing semantic appearance transfer across diverse image pairs, unlike GAN-based methods which are restricted to the dataset they have been trained on.

\subsection{Comparisons to Prior Work} \label{sec:comparison}

There are no existing methods that are tailored for solving our task:  semantic appearance transfer between two natural images (not restricted to a specific domain), without explicit user-guided inputs. We thus compare to prior works in which the problem setting is most similar to ours in some aspects (see discussion in these methods in Sec.~\ref{sec:related}):
(i) \emph{Swapping Autoencoders (SA)}~\cite{park2020swapping} -- a domain-specific, GAN-based method which has been trained to ``swap'' the texture and structure of two images in a realistic manner; (ii)  \emph{STROTSS}~\cite{kolkin2019style}, the style transfer method that also uses self-similarity of a pre-trained CNN features as the content descriptor, (ii) \emph{WCT$^2$}~\cite{Yoo_2019_ICCV}, a photorealistic NST method.

Since SA requires a dataset of images from two domains to train, we can only compare our results to their trained models on AHFQ and Flicker Mountain datasets.  For the rest of the methods, we also later compare to image pairs from our \emph{Wild-Pairs} examples. We evaluate our performance across a variety of image pairs both qualitatively, quantitatively and via an AMT user study. 

\begin{figure}
    \centering
    \includegraphics[width=.5\textwidth]{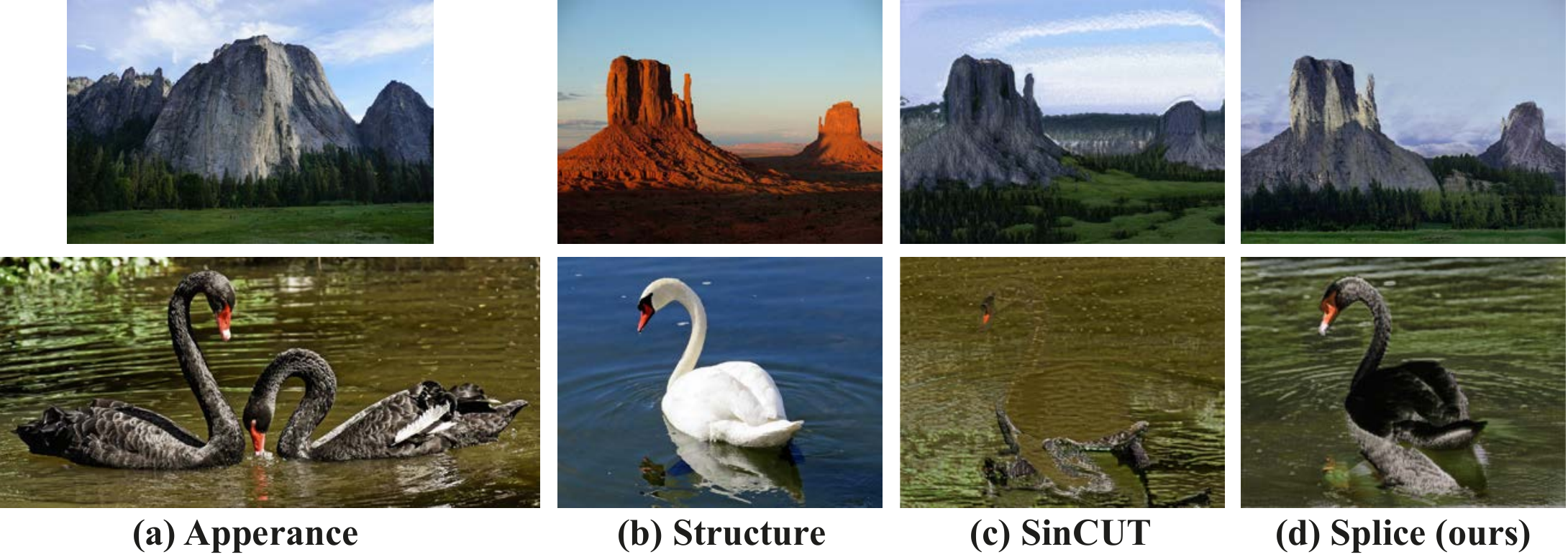}
    \caption{{\bf Comparison to SinCUT~\cite{park2020contrastive}.}  SinCUT results (c), when trained on each input pair (a-b), demonstrates it works well when the translation is mostly based on low-level information (top), but fails when higher level reasoning is required (bottom). (d) Our method successfully transfers the appearance across semantic regions, and generates high quality results w/o adversarial training.}\afterfigure
    \label{fig:sincut_comp}
\end{figure}

\subsubsection{Qualitative comparison}

Figure~\ref{fig:comparisons} shows sample results for all methods (additional results are included in the SM). In all examples, our method correctly relates semantically matching regions between the input images, and successfully transfer the visual appearance between them.  In the landscapes results (first 3 columns), it can be seen that SA outputs high quality images but  sometimes struggles to maintain high fidelity to the structure and appearance image: elements for the appearance image are often missing e.g., the fog in the left most example, or the trees in the second from left example. These visual elements are captured well in our results. For AHFQ, we noticed that SA often outputs a result that is nearly identical to the structure image. A possible cause to such behavior might be the adversarial loss, which ensures that the swapping result is a realistic image according to the the distribution of the training data. However, in some cases, this requirement does not hold (e.g. a German Shepherd with leopard's texture), and by outputting  the structure image the adversarial loss can be trivially satisfied.\footnote{We verified these results with the  authors \cite{park2020swapping}}.

NST frameworks such as STROTSS and WCT$^2$ well preserve the structure of the source image, but their results often depict visual artifacts:  STROTSS's results often suffer from color bleeding artifacts, while  WCT$^2$ results in  global color artifacts, demonstrating that transferring color is insufficient for tackling our task. 

Our method demonstrates better fidelity to the input structure and appearance images than GAN-based SA, while training only on the single input pair, without requiring a large collection of examples from each domain. With respect to style transfer, our method better transfers the appearance across semantically related regions in the input images, such as matching facial regions (e.g., eyes-to-eyes, nose-to-nose), while persevering the source structure. 

Finally, we also include a comparison to SinCUT~\cite{park2020contrastive}, a recent GAN-based image translation method. As demonstrated in Fig.~\ref{fig:sincut_comp},  SinCUT performs well for the landscape example, but since it can only utilize low-level visual information, it fails to  transfer the appearance of the swan in the second example. Our method successfully transfers the appearance across semantically realted regions, and generates high quality results w/o adversarial loss.

\begin{figure}[t!]
    \centering
    \includegraphics[width=.5\textwidth]{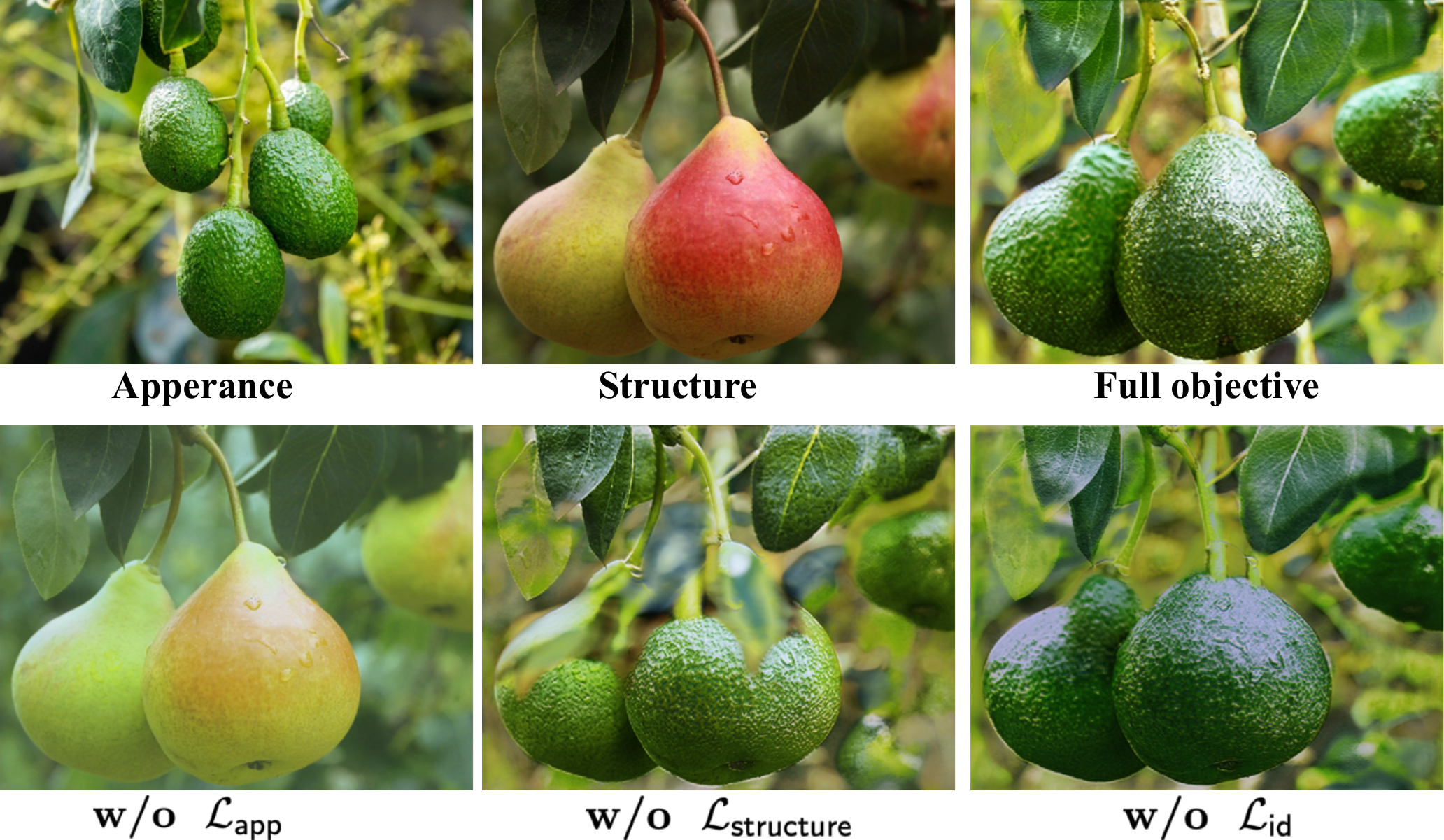}
    \caption{{\bf Loss ablation.} Our results when training without specific loss terms. When one of our loss terms is removed, the model fails to map the target appearance, preserve the input structure, or maintain fine details. See Sec.~\ref{sec:ablation} for more details.}\afterfigure
    \label{fig:ablation}
\end{figure}

\subsubsection{Quantitative comparison}
\label{sec:quantitive}

There is no existing automatic metric suitable for evaluating semantic appearance transfer across two natural images. We follow existing  style/appearance transfer methods, which mostly rely on human perceptual evaluation (e.g., \cite{JingYFYYS20, MechrezTZ18, kim2020deformable, park2020swapping}), and perform an extensive user study on Amazon Mechanical Turk (AMT). 

\myparagraph{Human Perceptual Evaluation} We design a user survey suitable for evaluating the task of  appearance transfer across semantically related scenes. We adopt the Two-alternative Forced Choice (2AFC) protocol suggested in \cite{park2020swapping, kolkin2019style}. Participants are shown with 2 reference images: the input structure image (A), shown in grayscale, and the input appearance image (B), along with 2 alternatives: our result and another baseline result. The participants are asked: \emph{``Which image best shows the shape/structure of~image~A combined with the appearance/style of image~B?''}.

We perform the survey using a collection of 65 images in total, gathered from AFHQ, Mountains, and Wild-Pairs. We collected 7000 user judgments w.r.t. existing baselines. Table \ref{table:amt} reports the percentage of votes in our favor.  As seen, our method outperforms all baselines across all image collections, especially for in the Wild-Pairs, which highlights our performance in challenging settings. Note that SA was trained on 500K mountain images, yet our method perform competitively. 
\begin{figure}[t!]
    \centering
    \includegraphics[width=.45\textwidth]{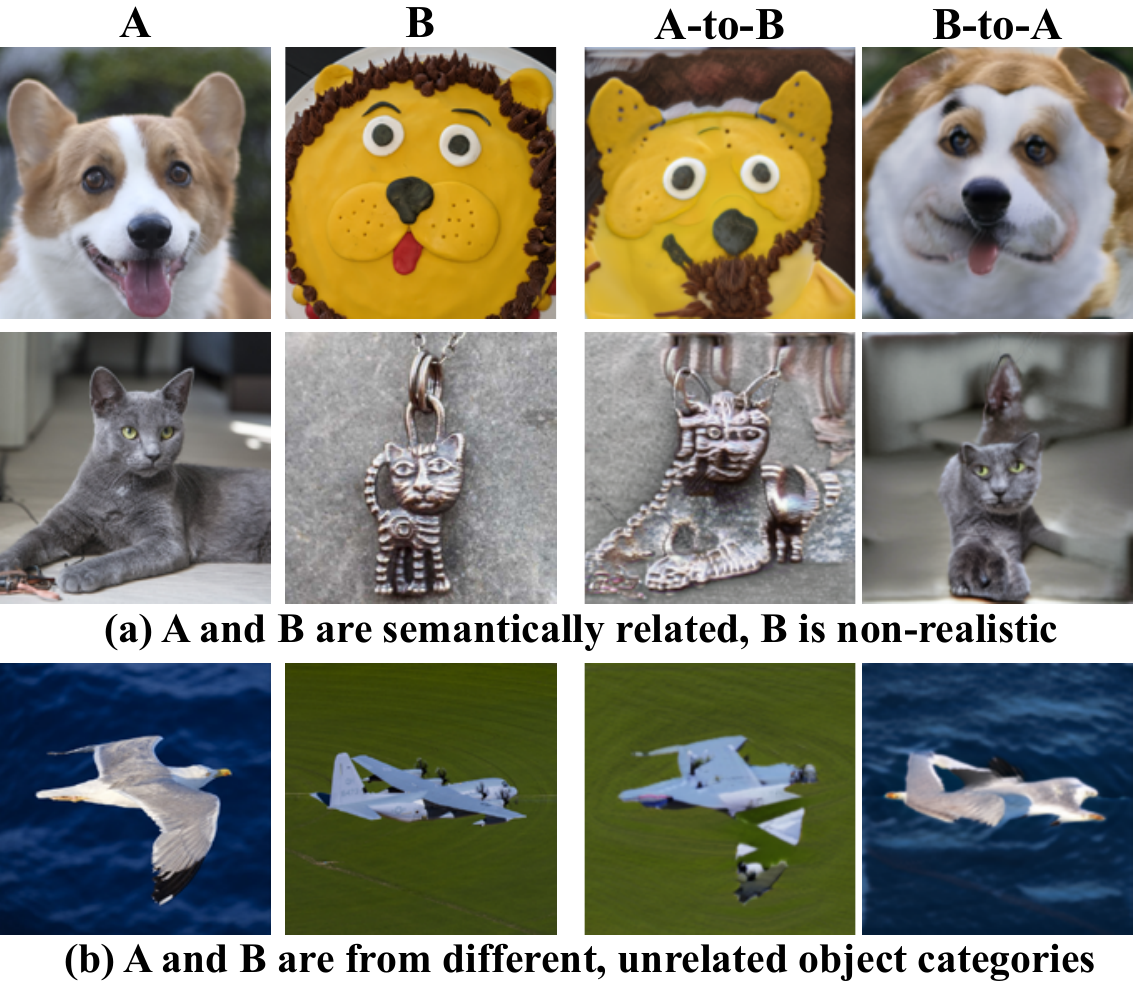}
    \caption{{\bf Semantic appearance transfer across different domains.} (a) Objects in the input images (A-B) are semantically related, yet B is non-realistic. (b) Objects are from unrelated object categories. See Sec.~\ref{sec:limitations} for discussion.}\afterfigure
    \label{fig:limitation}
\end{figure}

\begin{table} 
\centering
    \begin{adjustbox}{max width=\linewidth}
    \renewcommand{\tabcolsep}{4pt}
\begin{tabular}{c | c | c | c} 
 \hline
 & SA & STROTSS & WCT$^2$ \\ 
 \hline
 Wild-Pairs & - & \textbf{79.0} $\pm$ 13.0 & \textbf{83.1} $\pm$ 14.9 \\ 
 \hline
 mountains & \textbf{56.3} $\pm$ 10.0 & \textbf{58.8} $\pm$ 14.2 & \textbf{60.3} $\pm$ 12.1 \\
 \hline
 AFHQ & \textbf{71.8} $\pm$ 7.7 & \textbf{59.7} $\pm$ 15.3 & \textbf{61.0} $\pm$ 18.3 \\
 \hline
\end{tabular}
\end{adjustbox}
\label{table:amt}
\caption{\textbf{AMT perceptual evaluation}. We report results on AMT surveys evaluating the task of appearance transfer across semantically related scenes/objects (see Sec.~\ref{sec:quantitive}). For each dataset and a baseline, we report the percentage of judgments in our favor (mean, std). Our method outperforms   all  baselines: GAN-based, SA~\cite{park2020swapping}, and style transfer methods, STROTSS~\cite{kolkin2019style}, and WCT$^2$~\cite{Yoo_2019_ICCV}.}
\end{table}

\begin{table}[]
    \centering
    \renewcommand{\tabcolsep}{4pt}
    \begin{adjustbox}{max width=\linewidth}
    \begin{tabular}{c|c|c|c|c} 
    \hline
        & SA & STROTSS & WCT$^2$ & Splice (Ours) \\ 
        \hline
        Wild-Pairs & - & 0.83±0.11 & \textbf{0.89±0.06} & 0.88±0.06\\ 
        \hline
        mountains & 0.91±0.07 & 0.94±0.12 & \textbf{0.96±0.82} & 0.95±0.10\\
        \hline
    \end{tabular}
    \end{adjustbox}
    \caption{\textbf{mean IoU of output images with respect to the input structure images}. We extract semantic segmentation maps using Mask-RCNN \cite{he2017mask} for the Wild-Pairs collection, and \cite{zhou2018semantic_seg} for the mountains collection.}
    \label{tab:seg_results} \afterfigure
\end{table}

\myparagraph{Semantic layout preservation.} A key property of our method is the ability to preserve the semantic layout of the scene (while significantly changing the appearance of objects). We demonstrate this through the following evaluation.  We run semantic segmentation off-the-shelf model (e.g., MaskRCNN~\cite{he2017mask}) to compute object masks for the input structure images and  our results. Table \ref{tab:seg_results} reports IoU for our method and the baselines. Our method better preserves the scene layout than SA and STROTSS, and is the closet competitor to WCT$^2$ which only modifies colors, and as expected, achieves the highest IoU.

\subsection{Ablation }
\label{sec:ablation}
We  ablate the different loss terms in our objective by qualitatively comparing the results for our method when trained with our full objective (Eq.~\ref{eq:4}), and with a specific loss removed. The results are shown in Fig.~\ref{fig:ablation}. As can be seen, without the {\bf appearance loss} (w/o $\mathcal{L}_{\mathsf{app}}$), the model fails to map the target appearance, but only slightly modifies the colors of the input structure image due to the identity loss. That is, the identity loss encourages the model to learn an identity when it is fed with the target appearance image, and therefore even without the appearance loss some appearance supervision is available.
Without the {\bf structure loss}~(w/o $\mathcal{L}_{\mathsf{structure}}$), the model outputs an image with the desired appearance, but fails to fully preserve the structure of the input image, as can be seen by the distorted shape of the pears. Lastly, we observe that the {\bf identity loss} encourages the model to pay more attention to fine details both in terms of appearance and structure, e.g., the fine texture details of the avocado are refined.

\subsection{Limitations}
\label{sec:limitations}
Our performance depends on the internal representation learned by DINO-ViT. Therefore, in cases where the representation does not capture well the semantic association across objects in both images, our method would  fail to accomplish that too. Figure~\ref{fig:limitation} shows a few such cases: (a) objects are semantically related but one image is highly non-realistic (and thus out of distribution for \dinovit). For some regions, our methods successfully transfer the appearance but for some others it fails. In the cat example, we can see that in B-to-A result, the face and the body of the cat are  nicely mapped, yet our method fails to find a semantic correspondence for the rings, and we get a wrong mapping of the ear from image A.
In (b), our method does not manage to semantically relate a bird to an airplane. 

\section{Conclusions}
\label{sec:conclusions}

We  tackled a new problem setting in the context of style/appearance transfer:  semantically transferring appearance across related objects in two in-the-wild natural images, without any user guidance. Our approach demonstrates the power of DINO-ViT as an external semantic prior, and the effectiveness of utilizing it to establish or training losses -- we show how structure and appearance information can be disentangled from an input image, and then spliced together in a semantically meaningful way in the space of ViT features, through a generation process. 
We demonstrated that our method can be applied  on a variety of challenging input pairs across domains, in diverse poses and multiplicity of objects, and  can produce high-quality result without any adversarial training.  Our work unveils the potential of self-supervised representation learning not only for discriminative tasks such as image classification, but also for learning more powerful generative models. 

\paragraph{Acknowledgments:} We would like to thank Meirav Galun and Shir Amir for their insightful comments and discussion.
This project received funding from the Israeli Science Foundation (grant 2303/20), and the Carolito Stiftung. Dr Bagon is a Robin Chemers Neustein Artificial Intelligence Fellow.

{\small
\bibliographystyle{ieee_fullname}
\bibliography{egbib}
}

\appendix
\section{Implementation Details}\label{sec:sm_details}

\subsection{Generator Network Architecture}

We base our generator $G_\theta$ network on a \texttt{U-Net} architecture~\cite{ronneberger2015u}, with a 5-layer encoder and a symmetrical decoder. All layers comprise $3\!\times\!3$ Convolutions, followed by \texttt{BatchNorm}, and \texttt{LeakyReLU} activation. The encoder's channels dimensions are $[3\rightarrow16\rightarrow32\rightarrow64\rightarrow128\rightarrow128]$ (the decoder follows a reversed order).
In each level of the encoder, we add an additional $1\!\times\!1$ Convolution layer and concatenate the output features to the corresponding level of the decoder. Lastly, we add a $1\!\times\!1$ Convolution layer followed by \texttt{Sigmoid} activation to get the final RGB output.

\subsection{ViT Feature Extractor Architecture}
As described in Sec. 3, we leverage a pre-trained ViT model (\dinovit~\cite{dino}) trained in a self-supervised manner as a feature extractor. We use the 12 layer pretrained model in the $8\!\times\!8$ patches configuration (\texttt{ViT-B/8}), downloaded from the  \href{https://github.com/facebookresearch/dino}{official implementation at GitHub}.

\subsection{Training Details}
We implement our framework in PyTorch \cite{NEURIPS2019PyTorch} (code will be made available).  We optimize our full objective (Eq.~4, Sec.~3.3), with relative weights: $\alpha=0.1$, $\beta=0.1$. We use the Adam optimizer~\cite{DBLP:journals/corr/KingmaB14} with a constant learning rate of $\lambda=2\cdot 10^{-3}$. Each batch contains {$\{\tilde{I_s}, \tilde{I_t}\}$}, the augmented views of the source structure image and the target appearance image respectively. Every 75 iterations, we add {$\{I_s$, $I_t\}$} to the batch (i.e., do not apply augmentations). The resulting images  {$\{G(\tilde{I_s}), G(\tilde{I_t})\}$} and {$\tilde{I_t}$} are then resized down to $224$[pix] (maintaining aspect ratio) using bicubic interpolation, before extracting \dinovit features for estimating the losses. Training on an input image pair of size $512\!\times\!512$ takes $\sim\!20$ minutes to train on a single GPU (Nvidia RTX 6000) for a total of 2000 iterations.

\subsection{Data Augmentations}
\label{sec:appendix-aug}
We apply data augmentations to the input image pair {$\{I_s$, $I_t\}$} to create multiple \emph{internal examples} $\{I^i_s, I^i_t\}_{i=1}^N$. Specifically, at each training step, we apply the following augmentations:
\newline \newline Augmentations to the source structure image $I_s$:
\begin{itemize}
  \item random cropping: we uniformly sample a {NxN} crop such that N is between 95\% - 100\% of the height of $I_s$.
  \item random horizontal-flipping, applied in probability {p=0.5}.
  \item random color jittering: in probability {p=0.5} we jitter the brightness, contrast, saturation and hue of the image. 
  \item random Gaussian blurring: in probability {p=0.5} we apply a Gaussian blurring {3x3} filter ($\sigma$ is uniformly sampled between 0.1-2.0).
\end{itemize}
Augmentations to the target appearance image $I_t$:
\begin{itemize}
  \item random cropping: we uniformly sample a {NxN} crop such that N is between 95\% - 100\% of the height of $I_t$.
  \item random horizontal-flipping, applied in probability {p=0.5}.

\end{itemize}

\end{document}